\documentclass{Interspeech}

\interspeechcameraready

\title{LLM-Enhanced Dialogue Management for Full-Duplex Spoken Dialogue Systems
\thanks{\textit{*Corresponding author}}
}
\author[affiliation={1}]{Hao}{Zhang$^*$}
\author[affiliation={2}]{Weiwei}{Li}
\author[affiliation={3}]{Rilin}{Chen}
\author[affiliation={1}]{Vinay}{Kothapally}
\author[affiliation={1}]{Meng}{Yu}
\author[affiliation={1}]{Dong}{Yu}

\affiliation{Tencent AI Lab}{Bellevue}{USA}
\affiliation{Tencent AI Lab}{Shenzhen}{China}
\affiliation{Tencent AI Lab}{Beijing}{China}
\email{aaronhzhang@global.tentcent.com}

\usepackage{comment}
\usepackage{amsmath,multirow}
\usepackage[linesnumbered,ruled,vlined]{algorithm2e}
\usepackage{seqsplit}

\begin{document}
\maketitle

\begin{abstract}

Achieving full-duplex communication in spoken dialogue systems (SDS) requires real-time coordination between listening, speaking, and thinking. This paper proposes a semantic voice activity detection (VAD) module as a dialogue manager (DM) to efficiently manage turn-taking in full-duplex SDS. Implemented as a lightweight (0.5B) LLM fine-tuned on full-duplex conversation data, the semantic VAD predicts four control tokens to regulate turn-switching and turn-keeping, distinguishing between intentional and unintentional barge-ins while detecting query completion for handling user pauses and hesitations. By processing input speech in short intervals, the semantic VAD enables real-time decision-making, while the core dialogue engine (CDE) is only activated for response generation, reducing computational overhead. This design allows independent DM optimization without retraining the CDE, balancing interaction accuracy and inference efficiency for scalable, next-generation full-duplex SDS.

\end{abstract}

\section{Introduction}
Spoken dialogue systems (SDS) \cite{jokinen2009spoken, lemon2007machine, edlund2008towards, zhou2020design} have advanced significantly with large language models (LLMs), enabling more natural and context-aware interactions \cite{OpenAI2023ChatGPTSeeHearSpeak, OpenAI2023b}. However, achieving full-duplex communication, where SDS can listen and speak simultaneously, remains challenging \cite{hadi2023large, mctear2002spoken, zhou2023talking, mou2017media}. 
Many SDSs still operate in a turn-based (half-duplex) or ``pseudo'' full-duplex manner \cite{sacks1974simplest, skantze2021turn}, leading to less fluid interactions due to lack of understanding of the user's state and intention.
In contrast, human conversations involve seamless turn-taking \cite{zimmermann1996sex}. 
Beyond turn-taking, full-duplex SDS must handle challenges like interfering speakers, user hesitations, and distinguishing between intentional and unintentional interruptions to improve naturalness and efficiency \cite{liu2020towards, marge2022spoken, wang2021mell}.

Achieving full-duplex communication in SDS has been widely studied. Early approaches relied on separate neural network modules for detection and classification tasks, limiting stability and robustness. Shin et al. \cite{shin2024llm} used speech event detection for real-time query updates but lacked broader interaction handling. Lin et al. \cite{lin2022duplex} introduced Duplex Conversation, leveraging a multimodal model for user state detection and turn management. Recent methods embed interaction handling directly into LLMs, improving automation but increasing inference overhead. Wang et al. \cite{wang2024full} proposed NeuralFSM, fine-tuning an LLM with a finite state machine for synchronized speaking and listening. Similarly, Zhang et al. \cite{zhang2024beyond} fine-tuned an LLM with a time-division multiplexing strategy for duplex dialogue. VITA \cite{fu2024vita} incorporated state tokens and a duplex scheme for non-awakening and audio-interrupt interactions, while Moshi \cite{defossez2024moshi} integrated a base LLM with a smaller Transformer for real-time streaming predictions. Mai et al. \cite{mai2025real} introduced RTTL-DG, a textless spoken dialogue model incorporating backchannels and laughter for natural turn-taking.

To balance conversational quality, stability, and inference efficiency, this paper proposes a semantic voice activity detection (VAD) as a dialogue manager (DM) for full-duplex SDS. While an acoustic VAD mitigates interference from background speakers using acoustic cues, the semantic VAD leverages LLM capabilities to detect user states and intentions based on semantic information.
We implement the semantic VAD by fine-tuning a smaller (0.5B) LLM on carefully designed full-duplex text conversations. It predicts four control tokens—start-speaking, start-listening, continue-speaking, and continue-listening—to dynamically guide system interaction. Operating at short intervals, the DM enables real-time interaction management, distinguishing between intentional and unintentional barge-ins and detecting query completion to handle user pauses and hesitations. 
Our method improves upon prior work by balancing DM performance and computational efficiency: leveraging the semantic understanding of LLMs enable precise dialogue management, while a lightweight LLM cuts computational costs. Independent DM and CDE optimization further ensures scalability.

\section{Full-duplex spoken dialogue system}

\begin{figure*}[!t]
  \centering
  \includegraphics[width=1.0\linewidth]{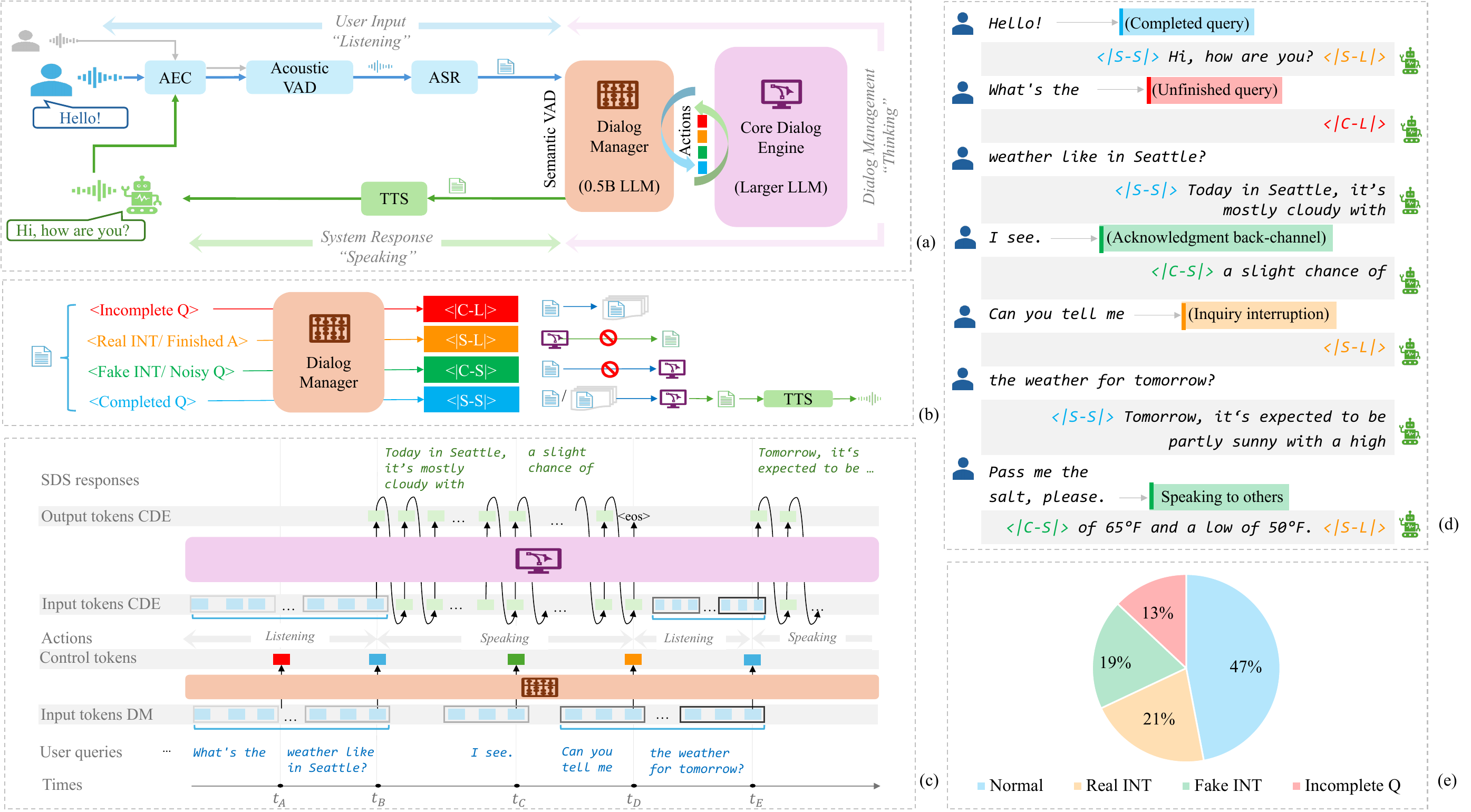}
  \caption{(a) The proposed system architecture; (b) Control tokens and their corresponding actions; (c) Interactions between the DM and CDE; (d) An example of a full-duplex conversation; and (e) Distribution of interaction scenarios in the generated full-duplex conversation dataset.}
  \label{fig:system}
\end{figure*}

\subsection{Key challenges in full-duplex SDS}

Achieving seamless full-duplex communication in SDS requires accurately detecting and differentiating user activities, states, and intentions. However, real-time turn-taking remains challenging due to issues such as interfering speakers, ambiguous pauses, and unintentional interruptions:

\begin{itemize}
    \item \textbf{Interfering Speakers}: Background speech can cause ASR and dialogue management errors, leading to unintended activations or responses.
    \item \textbf{User Pauses \& Hesitations}: Silence alone does not indicate query completion, often resulting in premature responses or unnecessary delays.
    \item \textbf{Unintentional Interruptions}: Acknowledgments, backchanneling, or speech directed at others may mistakenly halt SDS responses, disrupting conversational flow.
\end{itemize}
Addressing these challenges requires integrating both acoustic and semantic information for robust, context-aware full-duplex interaction.

\subsection{Proposed full-duplex SDS design}

The proposed SDS consists of six key modules: acoustic echo cancellation (AEC) \cite{zhang2018deep, zhang2022neural}, acoustic VAD, automatic speech recognition (ASR), semantic VAD, an LLM serving as the core dialogue engine (CDE), and text-to-speech (TTS), as shown in Figure~\ref{fig:system}(a). User speech is processed sequentially, with the first four modules operating in short intervals to ensure real-time responsiveness. The CDE is activated only when needed, minimizing computational overhead.

We combine acoustic and semantic VAD to handle the previously mentioned challenges, enabling robust and seamless full-duplex interaction:

\begin{itemize}
    \item \textbf{Acoustic VAD}: Our designed acoustic VAD is speaker- and distance-aware, leveraging distance information and speaker embeddings (if provided) to isolate target speaker’s activity and mitigate interference from other speakers.
    \item \textbf{Semantic VAD}: 
    Serving as the DM and implemented as a fine-tuned 0.5B LLM, it leverages the LLM's semantic understanding and contextual awareness to predict control tokens for dynamic turn-taking management. This enables effective handling of hesitations and accurate differentiation between real and fake interruptions.
\end{itemize}

This study focuses on introducing the semantic VAD.  Rather than improving question-answering capabilities, our goal is to train the DM to follow structured rules and take appropriate actions in various full-duplex interaction scenarios through instruction tuning \cite{ouyang2022training, wang2024full}. 
This approach requires only a small amount of high-quality data and minimal fine-tuning, making it efficient and practical. 
Unlike NN-based DMs that rely on shallow features \cite{shin2024llm, lin2022duplex}, the fine-tuned LLM leverages semantic understanding for precise interaction management. Compared to integrating DM directly into CDE (larger models) \cite{wang2024full, fu2024vita, zhang2024beyond, defossez2024moshi}, our method reduces inference costs by offloading frequent decisions to a lightweight LLM while ensuring scalability through independent DM and CDE optimization.

\section{Semantic VAD for dialogue management}

\subsection{Interaction scenarios and action tokens}

Most user-SDS interactions follow a standard question-answering flow without interruptions. However, when interruptions occur, we prioritize user experience in our design by allowing only user-initiated interruptions.

The DM enables SDS to switch between speaking and listening modes by performing two key tasks, each associated with specific action tokens (shown in Figure~\ref{fig:system}(b)):

\begin{itemize}
    \item \textbf{User State Detection}: Determines whether the user has finished speaking.
    \begin{itemize}
        \item \textit{Query Incomplete}: The DM predicts \texttt{\seqsplit{<|Continue-Listening|>}} (\texttt{<|C-L|>}), allowing the system to keep listening and cache historical queries until a complete query is identified.
        \item \textit{Query Complete}: The DM prompts the SDS to \texttt{\seqsplit{<|Start-Speaking|>}} (\texttt{<|S-S|>}). Once the response is finished, the SDS transitions to \texttt{\seqsplit{<|Start-Listening|>}} (\texttt{<|S-L|>}) again and awaits further input.
    \end{itemize}

    \item \textbf{User Intention Analysis}: Determines whether a user barge-in requires a response.
    \begin{itemize}
        \item \textit{Intentional Interruption (Real INT)}: If the user intends to redirect the conversation, the SDS stops speaking and switches to \texttt{<|Start-Listening|>} (\texttt{<|S-L|>}). Examples include denial or discontent, further inquiries, and topic changes.
        
        \item \textit{Unintentional Interruption (Fake INT)}: If the interruption is non-disruptive, the SDS continues speaking with \texttt{\seqsplit{<|Continue-Speaking|>}} (\texttt{<|C-S|>}). Examples include affirmative acknowledgments, back-channeling, speech directed at someone else, and unrelated comments.
    \end{itemize}
\end{itemize}

\begin{tiny} 
\begin{algorithm}[!t]
\caption{Full-Duplex Data Prompts Generation}
\label{alg:prompt_gen}
\SetKwInOut{Input}{Input}
\SetKwInOut{Output}{Output}
\Input{
    \textbf{Topic Pool} $T$; 
    \textbf{Speaking Style Pool} $S$; \\
    \textbf{Hyperparameters:} $N_{conv}$ (number of conversations), $P_{real}$ (probability of Real INT), $P_{fake}$ (Fake INT), $P_{incomplete}$ (incomplete Q)}
\Output{Generated prompts for full-duplex conversation data}

\For{$i \gets 1$ to $N_{conv}$}{
    Randomly select topic $t$ and speaking style $s$\;
    Set QA rounds $n \sim \text{Uniform}(2, 12)$\;
    Initialize conversation $C$ with $t$ and $s$\;

    \For{$j \gets 1$ to $n$}{
    Randomly assign QA type with probability $P_*$:
        \uIf{QA round is Real INT ($P_{real}$)}{
            Add real interruption scenario to $C$\;
        }
        \ElseIf{QA round is Fake INT ($P_{fake}$)}{
            Add fake interruption scenario to $C$\;
        }
        
        \Else{
            Add normal QA exchange to $C$\;
        }
        Randomly truncate user queries for incomplete queries ($P_{incomplete}$)\;
    }
    Generate a prompt by explaining control tokens placement and usage with examples\;
    Append prompt to prompt set\;
}
Return the set of generated prompts\;
\end{algorithm}
\end{tiny}

\subsection{Dialogue management}

As shown in Figure~\ref{fig:system}(c), the DM processes two primary inputs: (1) user query tokens from ASR and (2) historical response tokens from the CDE. For clarity and due to space constraints, the historical response tokens are not explicitly depicted in the diagram as DM input. The DM regulates interaction flow by analyzing these inputs to determine the next action.

The DM serves as the primary decision-maker, continuously managing listening and speaking states, while the CDE focuses on generating high-quality responses when required. The CDE is activated only when the DM predicts \texttt{<|S-S|>}, prompting it to perform standard LLM inference using past user queries and system responses.
By relying on the DM for frequent real-time decisions and selectively invoking the CDE, the system optimizes both computational efficiency and conversational quality, ensuring a scalable and effective full-duplex SDS.

\subsection{Full-duplex conversation data}
\label{Sect:data}

This section focuses on the core of the semantic VAD, detailing the data generation and the methodology for designing tailored prompts.

Creating high-quality training data for the dialogue manager (DM) requires annotated full-duplex conversations with the four control tokens, as shown in Figure~\ref{fig:system}(d). As no public datasets meet these criteria, we used LLM API, Yuanbao \cite{TencentChat2025}, to generate diverse and natural full-duplex dialogues.

\begin{table}[]
\centering
\caption{Confusion matrix full-duplex SDS evaluation. ``GT'' denotes ground truth, and ``Est'' denotes estimation.}
\resizebox{0.5\textwidth}{!}{%
\begin{tabular}{l|cccc|c}
\toprule
GT\textbackslash{}Est          & \texttt{<|C-L|>} & \texttt{<|S-S|>} & \texttt{<|S-L|>} & \texttt{<|C-S|>} & Recall   \\
\midrule
\texttt{<|C-L|>} & 926  & 74   & 0    & 0    & 0.926   \\
\texttt{<|S-S|>} & 11   & 989  & 0    & 0    & 0.989   \\
\texttt{<|S-L|>} & 1    & 0    & 999  & 0    & 0.999   \\
\texttt{<|C-S|>} & 0    & 0    & 0    & 1000 & 1.000   \\
\midrule
Precision        & 0.987 & 0.930 & 1.000 & 1.000 &Accuracy:  \\
F1 Score         & 0.956 & 0.959 & 0.999 & 1.000 & 0.9785 \\
\bottomrule
\end{tabular}
}
\label{table:1}
\end{table}

To generate prompts to create full duplex conversation data, we designed an algorithm as shown in Algorithm~\ref{alg:prompt_gen}.
The prompts were designed to instruct LLMs on how to generate full-duplex conversation data annotated with control tokens. Each prompt includes the following elements:
\begin{itemize}
    \item \textbf{Objective}: Generate a conversation transcript where the user can interrupt the assistant.
    \item \textbf{Assistant's Behavior}: Specify the meanings of control tokens and corresponding actions.
    \item \textbf{Output Format}: Conversations are formatted as:
        \texttt{\{Round X (dialogue type); User: <user's query or barge-in>; Sys: <assistant's response with control tokens>\}}.
    \item \textbf{Custom Instructions}: Specify the number of QA rounds, topic, and speaking style. For example:
        \texttt{"Please generate 5 rounds of conversation. The user initiates the conversation with a topic related to travel, speaking in a casual and conversational style. The assistant maintains a friendly tone and references the previous context."}
\end{itemize}

The topic pool covers 200 common conversation themes (e.g., weather, travel, restaurants), while the speaking style pool includes 10 user personas. Topics and styles are randomly selected, with probabilities adjusted for realism. 

\begin{table*}[!t]
\centering
\caption{Comparison with related studies.}
\label{table:2}
\resizebox{\textwidth}{!}{%
\begin{tabular}{lll|ccc|ccc|ccc}
\toprule
\multicolumn{3}{c|}{\textbf{Tasks \& Actions} }                                        & \multicolumn{3}{c|}{\textbf{Precision}}                                             & \multicolumn{3}{c|}{\textbf{Recall}}                                                  & \multicolumn{3}{c}{\textbf{F1}}                                                                  \\ \hline
\multicolumn{1}{c|}{\begin{tabular}[c]{@{}c@{}}DuplexConv\\ {\cite{lin2022duplex}}\end{tabular}}                     & \multicolumn{1}{c|}{\begin{tabular}[c]{@{}c@{}}RTTL-DG\\ {\cite{mai2025real}}\end{tabular}}    & \multicolumn{1}{c|}{\begin{tabular}[c]{@{}c@{}}SemanticVAD\\ (proposed)\end{tabular}} & \multicolumn{1}{c|}{{\cite{lin2022duplex}}}               & \multicolumn{1}{c|}{{\cite{mai2025real}}} & SemanticVAD & \multicolumn{1}{c|}{{\cite{lin2022duplex}}}               & \multicolumn{1}{c|}{{\cite{mai2025real}}} & SemanticVAD & \multicolumn{1}{c|}{{\cite{lin2022duplex}}}               & \multicolumn{1}{c|}{{\cite{mai2025real}}} & SemanticVAD \\ \hline
\multicolumn{1}{l|}{\multirow{2}{*}{\begin{tabular}[c]{@{}l@{}}User state \\ detection\end{tabular}}} & \multicolumn{1}{l|}{\begin{tabular}[c]{@{}l@{}}   Remain \\ silent \end{tabular}} & \begin{tabular}[c]{@{}l@{}}    Continue listening \\ \texttt{<|C-L|>}      \end{tabular}                         & \multicolumn{1}{c|}{\multirow{2}{*}{/}}    & \multicolumn{1}{c|}{0.81}    & 0.99        & \multicolumn{1}{c|}{\multirow{2}{*}{/}}    & \multicolumn{1}{c|}{0.89}    & 0.93        & \multicolumn{1}{c|}{\multirow{2}{*}{0.91}} & \multicolumn{1}{c|}{0.85}    & 0.96        \\ \cline{2-3} \cline{5-6} \cline{8-9} \cline{11-12} 
\multicolumn{1}{l|}{}                                                                                 & \multicolumn{1}{l|}{\begin{tabular}[c]{@{}l@{}}Initiate  \\  speaking\end{tabular}}     & \begin{tabular}[c]{@{}l@{}}Start speaking\\ \texttt{<|S-S|>}\end{tabular}                      & \multicolumn{1}{c|}{}                      & \multicolumn{1}{c|}{0.62}    & 0.93        & \multicolumn{1}{c|}{}                      & \multicolumn{1}{c|}{0.43}    & 0.99        & \multicolumn{1}{c|}{}                      & \multicolumn{1}{c|}{0.52}    & 0.96        \\ \hline
\multicolumn{1}{l|}{\multirow{2}{*}{\begin{tabular}[c]{@{}l@{}}User intention \\ detection\end{tabular}}}   & \multicolumn{1}{l|}{\begin{tabular}[c]{@{}l@{}}Stop \\ speaking\end{tabular}}     & \begin{tabular}[c]{@{}l@{}}Start listening \\ \texttt{<|S-L|>}  \end{tabular}                        & \multicolumn{1}{c|}{\multirow{2}{*}{0.91}} & \multicolumn{1}{c|}{0.75}    & 0.99        & \multicolumn{1}{c|}{\multirow{2}{*}{0.86}} & \multicolumn{1}{c|}{0.53}    & 0.93        & \multicolumn{1}{c|}{\multirow{2}{*}{0.89}} & \multicolumn{1}{c|}{0.62}    & 0.96        \\ \cline{2-3} \cline{5-6} \cline{8-9} \cline{11-12} 
\multicolumn{1}{l|}{}                                                                                 & \multicolumn{1}{l|}{\begin{tabular}[c]{@{}l@{}}Keep \\ speaking\end{tabular}}     & \begin{tabular}[c]{@{}l@{}}Continue speaking\\ \texttt{<|C-S|>}\end{tabular}                       & \multicolumn{1}{c|}{}                      & \multicolumn{1}{c|}{0.94}    & 1.00        & \multicolumn{1}{c|}{}                      & \multicolumn{1}{c|}{0.96}    & 1.00        & \multicolumn{1}{c|}{}                      & \multicolumn{1}{c|}{0.95}    & 1.00        \\ \bottomrule
\end{tabular}
}
\end{table*}

\begin{table*}[!t]
\centering
\caption{Testing using real recorded data with manually labeled user query completion.}
\label{table:3}
\resizebox{0.8\textwidth}{!}{%
\begin{tabular}{c|c|cc|c|cc|c|c|c|c}
\toprule
                        &                                & \multicolumn{2}{c|}{\textbf{AcousticVAD}}                                 & \multirow{10}{*}{$\Rightarrow$} & \multicolumn{6}{c}{\textbf{+ SemanticVAD}}                       \\ \cline{1-4} \cline{6-11}
Threshold               & GT\textbackslash{}Est          & \texttt{<|C-L|>} & \texttt{<|S-S|>} &                      & \texttt{<|C-L|>} & \texttt{<|S-S|>} & Recall & Precision & F1    & Accuracy               \\ \cline{1-4} \cline{6-11}
\multirow{2}{*}{300 ms}  & \texttt{<|C-L|>} & 0                              & 532                            &                      & 495                            & 37                             & 0.930  & 0.887     & 0.908 & \multirow{2}{*}{0.935} \\
                        & \texttt{<|S-S|>} & 0                              & 1008                           &                      & 63                             & 945                            & 0.937  & 0.962     & 0.949 &                        \\ \cline{1-4} \cline{6-11}
\multirow{2}{*}{500 ms}  & \texttt{<|C-L|>} & 0                              & 324                            &                      & 307                            & 17                             & 0.949  & 0.892     & 0.920 & \multirow{2}{*}{0.962} \\
                        & \texttt{<|S-S|>} & 0                              & 1078                           &                      & 37                             & 1041                           & 0.966  & 0.984     & 0.975 &                        \\ \cline{1-4} \cline{6-11}
\multirow{2}{*}{800 ms}  & \texttt{<|C-L|>} & 0                              & 228                            &                      & 218                            & 10                             & 0.956  & 0.861     & 0.905 & \multirow{2}{*}{0.966} \\
                        & \texttt{<|S-S|>} & 0                              & 1091                           &                      & 35                             & 1056                           & 0.968  & 0.991     & 0.979 &                        \\ \cline{1-4} \cline{6-11}
\multirow{2}{*}{1800 ms} & \texttt{<|C-L|>} & 0                              & 132                            &                      & 128                            & 4                              & 0.970  & 0.800     & 0.880 & \multirow{2}{*}{0.971} \\
                        & \texttt{<|S-S|>} & 0                              & 1108                           &                      & 32                             & 1076                           & 0.971  & 0.996     & 0.983 &    \\
                        \bottomrule
\end{tabular}
}
\label{tab:realdata}
\end{table*}

\subsection{Data preparation and model training}

Data preparation consists of four stages: data generation, cleaning, post-processing, and augmentation \footnote{The data preparation scripts are available at: \url{https://github.com/HaoZhang6720/fullduplex-dialogue-data}}.
We generated full-duplex conversations using 20,000 prompts with Yuanbao \cite{TencentChat2025}, but found that not all generated data align fully with the intended structure due to limited command-following abilities. To address this, we applied strict data cleaning, filtering malformed outputs, and incorrect control tokens, resulting in a retention rate of 60\%. 
Given this, we also adopted an alternative strategy: first generating standard QA dialogues, which LLMs handle more effectively, and then introduce various interaction patterns via controlled post-processing. The final training dataset combines conversations from both strategies.

We further augmented the data by modifying the punctuation at the end of user queries to better match the ASR output. The final dataset includes 11,990 conversations with 80,338 dialogue rounds covering diverse interaction scenarios (Figure~\ref{fig:system}(e)). Furthermore, we examine extreme cases with a high proportion (over 50\%) or exclusively real/fake interruptions or incomplete queries, which led to overfitting and degraded performance, reinforcing the need for balanced training data. To mitigate potential degradation of the base LLM’s capabilities,
we included the corresponding uninterrupted dialogues alongside full-duplex samples in the training set.

For fine-tuning, we used a small version of Hunyuan \cite{sun2024hunyuan}, 0.5B-dense-8k model, on the curated dataset. The initial study focuses on Chinese dialogues and the dataset exclusively consists of Chinese full-duplex dialogues. The training follows standard LLM fine-tuning with four added control tokens, running for 1500 steps with a batch size of 128 and a learning rate of 0.001, linearly decayed to 0.0001 for stable optimization.

\section{Experimental results}

\subsection{Test sets}
As no benchmark datasets exist for full-duplex evaluation, we generate 1,000 test samples for each interaction scenario.
Following Sect.~\ref{Sect:data}, we reconstruct the topic and speaking style pools, and generate 2,000 multi-round conversations. From these, 1,000 samples per scenario are randomly selected to evaluate the semantic VAD’s control token predictions.

\subsection{Evaluation results}
The evaluation results presented in Table~\ref{table:1} confirm the effectiveness of the proposed semantic VAD in predicting control tokens across all scenarios. 
Notably, the detection performance for user barge-in (determining whether the system should continue speaking or switch to start listening) is exceptionally high, benefiting from contextual cues during simultaneous speech, making it more stable.
In contrast, user state detection (distinguishing between finished and unfinished queries) is slightly lower as it relies solely on semantic completeness, which varies with speaking style and linguistic nuances, introducing ambiguity and making it inherently more challenging for the proposed method.

Table~\ref{table:2} compares our approach with two recently proposed full-duplex SDS methods \cite{lin2022duplex,mai2025real}. As we do not have access to their model checkpoints or test datasets, we can only present the results reported in their studies directly for reference. Both studies framed user state and barge-in detection as classification tasks based on acoustic and linguistic patterns with limited semantic understanding. In contrast, our semantic VAD leverages LLM-driven semantic comprehension, enabling more context-aware and reliable full-duplex interaction, achieving relatively better performance.

\subsection{Evaluation on real-recordings}

We further tested our system using internal recordings from user interactions with a half-duplex SDS, focusing on user state detection (start-speaking and continue-listening). Users were instructed to include natural hesitations to simulate scenarios with incomplete queries. 
The recordings were labeled using both VAD techniques and manual annotation for evaluation.
Table~\ref{table:3} presents the results. Since acoustic VAD relies solely on acoustic information and determines the completion of the query by comparing the silence duration against a fixed threshold (e.g. 300 ms), it can only predict \texttt{<|S-S|>}—assuming the user has finished speaking. 
Semantic VAD refinement significantly improved accuracy, achieving over 93.5\% across all cases. Further error analysis indicated that some mispredictions were introduced due to ASR errors rather than limitations of the VAD itself.

\section{Conclusion}

This paper introduces a semantic VAD-based dialogue manager (DM) for full-duplex SDS, leveraging a fine-tuned 0.5B LLM to regulate turn-taking through control tokens. The proposed approach effectively distinguishes between intentional and unintentional barge-ins, detects query completion, and reduces computational overhead by selectively invoking the core dialogue engine (CDE). Experimental results demonstrate improved interaction fluidity and intent recognition. Future work will address system delay and robustness while extending this method to support large multimodal models.

\bibliographystyle{IEEEtran}
\bibliography{mybib}

@inproceedings{lin2022duplex,
  title={Duplex conversation: Towards human-like interaction in spoken dialogue systems},
  author={Lin, Ting-En and Wu, Yuchuan and Huang, Fei and Si, Luo and Sun, Jian and Li, Yongbin},
  booktitle={Proceedings of the 28th ACM SIGKDD Conference on Knowledge Discovery and Data Mining},
  pages={3299--3308},
  year={2022}
}

@article{shin2024llm,
  title={LLM-based Natural Conversational Agent with Speech Collision Detection for Early Prompt Abort},
  author={Shin, D},
  year={2024}
}

@article{wang2024full,
  title={A full-duplex speech dialogue scheme based on large language models},
  author={Wang, Peng and Lu, Songshuo and Tang, Yaohua and Yan, Sijie and Xia, Wei and Xiong, Yuanjun},
  journal={arXiv preprint arXiv:2405.19487},
  year={2024}
}

@article{zhang2024beyond,
  title={Beyond the turn-based game: Enabling real-time conversations with duplex models},
  author={Zhang, Xinrong and Chen, Yingfa and Hu, Shengding and Han, Xu and Xu, Zihang and Xu, Yuanwei and Zhao, Weilin and Sun, Maosong and Liu, Zhiyuan},
  journal={arXiv preprint arXiv:2406.15718},
  year={2024}
}

@article{fu2024vita,
  title={Vita: Towards open-source interactive omni multimodal llm},
  author={Fu, Chaoyou and Lin, Haojia and Long, Zuwei and Shen, Yunhang and Zhao, Meng and Zhang, Yifan and Dong, Shaoqi and Wang, Xiong and Yin, Di and Ma, Long and others},
  journal={arXiv preprint arXiv:2408.05211},
  year={2024}
}

@article{defossez2024moshi,
  title={Moshi: a speech-text foundation model for real-time dialogue},
  author={D{\'e}fossez, Alexandre and Mazar{\'e}, Laurent and Orsini, Manu and Royer, Am{\'e}lie and P{\'e}rez, Patrick and J{\'e}gou, Herv{\'e} and Grave, Edouard and Zeghidour, Neil},
  journal={arXiv preprint arXiv:2410.00037},
  year={2024}
}

@article{mai2025real,
  title={Real-Time Textless Dialogue Generation},
  author={Mai, Long and Carson-Berndsen, Julie},
  journal={arXiv preprint arXiv:2501.04877},
  year={2025}
}

@book{jokinen2009spoken,
  title={Spoken dialogue systems},
  author={Jokinen, Kristina and McTear, Michael},
  year={2009},
  publisher={Morgan \& Claypool Publishers}
}

@inproceedings{lemon2007machine,
  title={Machine learning for spoken dialogue systems},
  author={Lemon, Olivier and Pietquin, Olivier},
  booktitle={European Conference on Speech Communication and Technologies (Interspeech'07)},
  pages={2685--2688},
  year={2007}
}

@article{edlund2008towards,
  title={Towards human-like spoken dialogue systems},
  author={Edlund, Jens and Gustafson, Joakim and Heldner, Mattias and Hjalmarsson, Anna},
  journal={Speech communication},
  volume={50},
  number={8-9},
  pages={630--645},
  year={2008},
  publisher={Elsevier}
}

@misc{OpenAI2023ChatGPTSeeHearSpeak,
  title        = {{ChatGPT can now see, hear, and speak}},
  author       = {{OpenAI}},
  year         = {2023},
  howpublished = {\url{https://openai.com/blog/chatgpt-can-now-see-hear-and-speak.}},
  note         = {[Online; accessed 2025-02-11]}
}

@misc{OpenAI2023b,
  author       = {{OpenAI}},
  title        = {{Introducing ChatGPT}},
  year         = {2023},
  howpublished = {\url{https://openai.com/blog/chatgpt#OpenAI}},
  note         = {[Online; accessed 2025-02-11]}
}

@article{mctear2002spoken,
  title={Spoken dialogue technology: enabling the conversational user interface},
  author={McTear, Michael F},
  journal={ACM Computing Surveys (CSUR)},
  volume={34},
  number={1},
  pages={90--169},
  year={2002},
  publisher={ACM New York, NY, USA}
}

@article{zhou2023talking,
  title={Talking to a bot or a wall? How chatbots vs. human agents affect anticipated communication quality},
  author={Zhou, Qi and Li, Bin and Han, Lei and Jou, Min},
  journal={Computers in Human Behavior},
  volume={143},
  pages={107674},
  year={2023},
  publisher={Elsevier}
}

@article{mou2017media,
  title={The media inequality: Comparing the initial human-human and human-AI social interactions},
  author={Mou, Yi and Xu, Kun},
  journal={Computers in Human Behavior},
  volume={72},
  pages={432--440},
  year={2017},
  publisher={Elsevier}
}

@article{skantze2021turn,
  title={Turn-taking in conversational systems and human-robot interaction: a review},
  author={Skantze, Gabriel},
  journal={Computer Speech \& Language},
  volume={67},
  pages={101178},
  year={2021},
  publisher={Elsevier}
}

@incollection{zimmermann1996sex,
  title={Sex roles, interruptions and silences in conversation},
  author={Zimmermann, Don H and West, Candace},
  booktitle={Amsterdam Studies in the Theory and History of Linguistic Science Series 4},
  pages={211--236},
  year={1996},
  publisher={John Benjamins BV}
}

@article{hadi2023large,
  title={Large language models: a comprehensive survey of its applications, challenges, limitations, and future prospects},
  author={Hadi, Muhammad Usman and Qureshi, Rizwan and Shah, Abbas and Irfan, Muhammad and Zafar, Anas and Shaikh, Muhammad Bilal and Akhtar, Naveed and Wu, Jia and Mirjalili, Seyedali and others},
  journal={Authorea Preprints},
  year={2023},
  publisher={Authorea}
}

@inproceedings{liu2020towards,
  title={Towards building an intelligent chatbot for customer service: Learning to respond at the appropriate time},
  author={Liu, Che and Jiang, Junfeng and Xiong, Chao and Yang, Yi and Ye, Jieping},
  booktitle={Proceedings of the 26th ACM SIGKDD international conference on Knowledge Discovery \& Data Mining},
  pages={3377--3385},
  year={2020}
}

@article{marge2022spoken,
  title={Spoken language interaction with robots: Recommendations for future research},
  author={Marge, Matthew and Espy-Wilson, Carol and Ward, Nigel G and Alwan, Abeer and Artzi, Yoav and Bansal, Mohit and Blankenship, Gil and Chai, Joyce and Daum{\'e} III, Hal and Dey, Debadeepta and others},
  journal={Computer Speech \& Language},
  volume={71},
  pages={101255},
  year={2022},
  publisher={Elsevier}
}

@inproceedings{wang2021mell,
  title={Mell: Large-scale extensible user intent classification for dialogue systems with meta lifelong learning},
  author={Wang, Chengyu and Pan, Haojie and Liu, Yuan and Chen, Kehan and Qiu, Minghui and Zhou, Wei and Huang, Jun and Chen, Haiqing and Lin, Wei and Cai, Deng},
  booktitle={Proceedings of the 27th ACM SIGKDD conference on knowledge discovery \& data mining},
  pages={3649--3659},
  year={2021}
}

@article{sacks1974simplest,
  title={A simplest systematics for the organization of turn-taking for conversation},
  author={Sacks, Harvey and Schegloff, Emanuel A and Jefferson, Gail},
  journal={language},
  volume={50},
  number={4},
  pages={696--735},
  year={1974},
  publisher={Linguistic Society of America}
}

@article{zhou2020design,
  title={The design and implementation of xiaoice, an empathetic social chatbot},
  author={Zhou, Li and Gao, Jianfeng and Li, Di and Shum, Heung-Yeung},
  journal={Computational Linguistics},
  volume={46},
  number={1},
  pages={53--93},
  year={2020},
  publisher={MIT Press One Rogers Street, Cambridge, MA 02142-1209, USA journals-info~…}
}

@article{ouyang2022training,
  title={Training language models to follow instructions with human feedback},
  author={Ouyang, Long and Wu, Jeffrey and Jiang, Xu and Almeida, Diogo and Wainwright, Carroll and Mishkin, Pamela and Zhang, Chong and Agarwal, Sandhini and Slama, Katarina and Ray, Alex and others},
  journal={Advances in neural information processing systems},
  volume={35},
  pages={27730--27744},
  year={2022}
}

@article{zhang2018deep,
  title={Deep learning for acoustic echo cancellation in noisy and double-talk scenarios},
  author={Zhang, Hao and Wang, D},
  journal={Training},
  volume={161},
  number={2},
  pages={322},
  year={2018}
}

@article{zhang2022neural,
  title={Neural cascade architecture for multi-channel acoustic echo suppression},
  author={Zhang, Hao and Wang, DeLiang},
  journal={IEEE/ACM Transactions on Audio, Speech, and Language Processing},
  volume={30},
  pages={2326--2336},
  year={2022},
  publisher={IEEE}
}

@misc{TencentChat2025,
  author       = {{Yuanbao}},
  title        = {Tencent Yuanbao Chat},
  year         = {2025},
  howpublished = {\url{https://yuanbao.tencent.com/chat/}},
  note         = {[Online; accessed 2025-02-11]}
}

@article{sun2024hunyuan,
  title={Hunyuan-large: An open-source moe model with 52 billion activated parameters by tencent},
  author={Sun, Xingwu and Chen, Yanfeng and Huang, Yiqing and Xie, Ruobing and Zhu, Jiaqi and Zhang, Kai and Li, Shuaipeng and Yang, Zhen and Han, Jonny and Shu, Xiaobo and others},
  journal={arXiv preprint arXiv:2411.02265},
  year={2024}
}

\end{document}